\documentclass[runningheads]{llncs}
\usepackage{graphicx}
\usepackage{subfigure}
\usepackage{amsmath,amssymb} 
\usepackage{color}
\usepackage{url}
\usepackage[pagebackref=true,breaklinks=true,letterpaper=true,colorlinks,citecolor=blue,linkcolor=blue,bookmarks=false]{hyperref}

\usepackage{multirow}
\usepackage{booktabs} 
\usepackage{array}

\newcolumntype{P}[1]{>{\centering\arraybackslash}p{#1}}

\begin{document}
\title{PSDF Fusion: Probabilistic Signed Distance Function for On-the-fly 3D Data Fusion and Scene Reconstruction} 

\titlerunning{PSDF Fusion}
%
\author{Wei Dong \and
Qiuyuan Wang \and
Xin Wang \and
Hongbin Zha}
%
\authorrunning{Wei Dong, Qiuyuan Wang, Xin Wang, and Hongbin Zha}
%

\institute{Key Laboratory of Machine Perception (MOE), School of EECS, Peking University \\
Cooperative Medianet Innovation Center, Shanghai Jiao Tong University \\
\email{\{w.dong,qiuyuanwang,xinwang\_cis\}@pku.edu.cn, zha@cis.pku.edu.cn}}
\maketitle              
%
\begin{abstract}
	We propose a novel 3D spatial representation for data fusion and scene reconstruction. 
	\textit{Probabilistic Signed Distance Function} (Probabilistic SDF, PSDF) is proposed to depict uncertainties in the 3D space. It is modeled by a joint distribution describing SDF value and its inlier probability,  reflecting input data quality and surface geometry. 
	A \textit{hybrid data structure} involving voxel, surfel, and mesh is designed to fully exploit the advantages of various prevalent 3D representations. Connected by PSDF, these components reasonably cooperate in a consistent framework.	
	Given sequential depth measurements, PSDF can be incrementally refined with less ad hoc parametric Bayesian updating.
	Supported by PSDF and the efficient 3D data representation, high-quality surfaces can be extracted on-the-fly, and in return contribute to reliable data fusion using the geometry information.
	Experiments demonstrate that our system reconstructs scenes with higher model quality and lower redundancy, and runs faster than existing online mesh generation systems.
\end{abstract}

\section{Introduction}		
\begin{figure}
	\centering
	\includegraphics[width=0.9\textwidth]{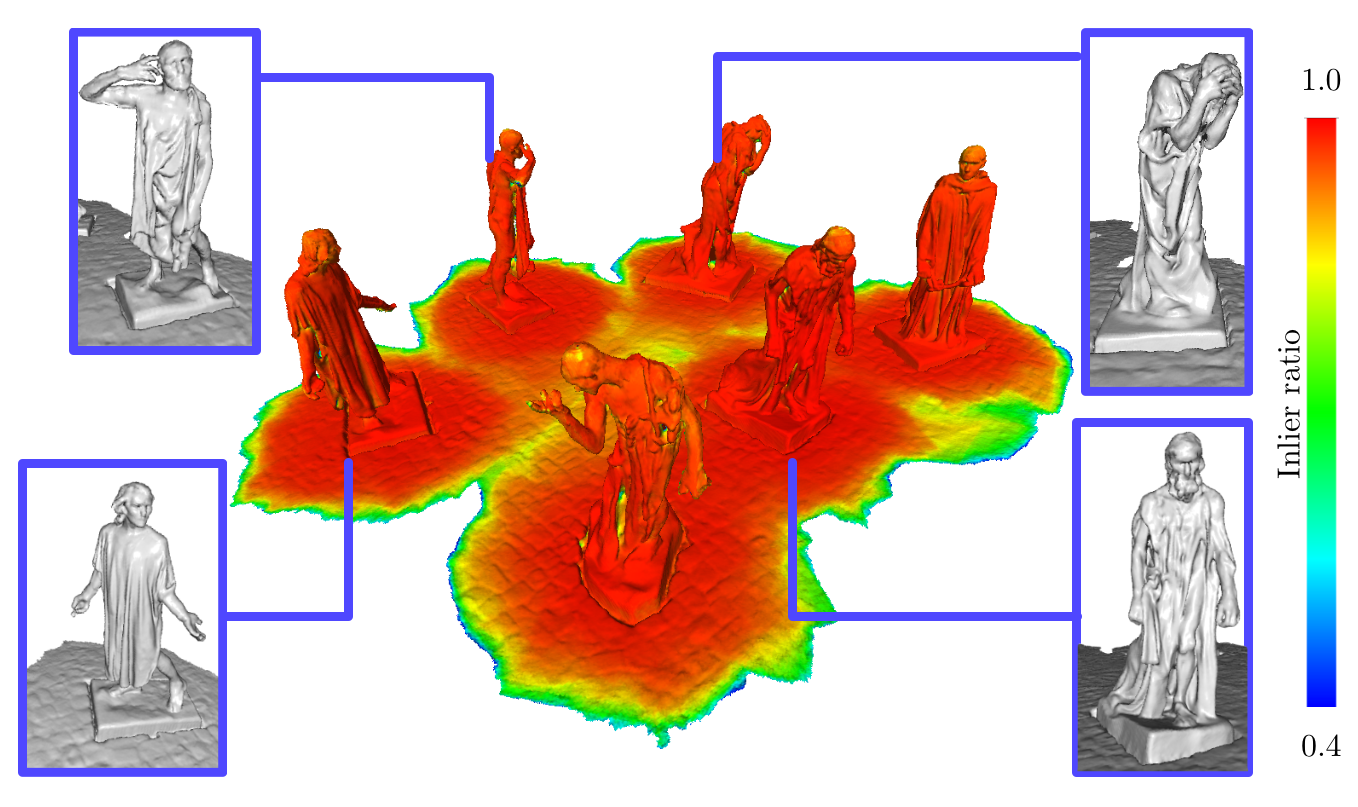}
	\caption{Reconstructed mesh of {\it burghers}. The heatmap denotes the SDF inlier ratio (SDF confidence). Note details are preserved and outliers have been all removed without any post-processing. Inlier-ratio heatmaps in Fig.\ref{fig:tum3},\ref{fig:copyroom+lounge},\ref{fig:burghers-sequential} also conform to this colorbar.}\label{fig:burghers-teaser}
\end{figure}

In recent years, we have witnessed the appearance of consumer-level depth sensors and the increasing demand of real-time 3D geometry information in next-generation applications. Therefore, online dense scene reconstruction has been a popular research topic. The essence of the problem is to fuse noisy depth data stream into a reliable 3D representation where clean models can be extracted. It is necessary to consider uncertainty in terms of sampling density, measurement accuracy, and surface complexity so as to better understand the 3D space.

Many representations built upon appropriate mathematical models are designed for robust data fusion in such a context. To handle uncertainties, {\it surfel} and {\it point} based approaches \cite{vogiatzis2011video,Keller2013,Kolev2014} adopt filtering-based probabilistic models that explicitly manipulate input data. {\it Volume} based methods~\cite{Newcombe2011,kahler2015very,dai2017bundlefusion,Ulusoy:CVPR:2016}, on the other hand, maximize spatial probabilistic distributions and output discretized 3D properties such as SDF and occupancy state. With fixed topologies, \textit{mesh} based methods~\cite{Zienkiewicz2016} may also involve parametric minimization of error functions.

While such representations have been proven effective by various applications, their underlying data structures endure more or less drawbacks. Surfels and points are often loosely managed without topology connections, requiring additional modules for efficient indexing and rendering~\cite{Botsch2003}, and is relatively prone to noisy input. Volumetric grids lack flexibility to some extent, hence corresponding data fusion can be either oversimplified using weighted average~\cite{Newcombe2011}, or much time-consuming in order to maximize joint distributions~\cite{Ulusoy:CVPR:2016}. In addition, ray-casting based volume rendering is also non-trivial. Usually storing vertices with strong topological constraints, mesh is similarly hard to manipulate and is less applicable to many situations. There have been studies incorporating aforementioned data structures \cite{Marton2009,Steinbrucker2014,Klingensmith2015,InfiniTAM_V3_Report_2017}, yet most of these pipelines are loosely organized without fully taking the advantages of each representation.

In this paper, we design a novel framework to fully exploit the power of existing 3D representations, supported by PSDF-based probabilistic computations. Our framework is able to perform reliable depth data fusion and reconstruct high-quality surfaces in real-time with more details and less noise, as depicted in Fig.\ref{fig:burghers-teaser}. Our contributions can be concluded as:
\begin{enumerate}
\item We present a novel hybrid data structure integrating voxel, surfel, and mesh;
\item The involved 3D representations are systematically incorporated in the consistent probabilistic framework linked by the proposed PSDF;
\item Incremental 3D data fusion is built upon less ad-hoc probabilistic computations in a parametric Bayesian updating fashion, contributes to online surface reconstruction, and benefits from iteratively recovered geometry in return.
\end{enumerate}
\section{Related Work}
\noindent \textbf{Dense reconstruction from depth images.}
There is a plethora of successful off-the-shelf systems that reconstruct 3D scenes from depth scanners' data.
\cite{Curless1996} presents the volumetric 3D grids to fuse multi-view range images. \cite{Newcombe2011} extends \cite{Curless1996} and proposes KinectFusion, a real-time tracking and dense mapping system fed by depth stream from a consumer-level Kinect. KinectFusion has been improved by VoxelHashing \cite{Niessner2013}, InfiniTAM \cite{kahler2015very}, and BundleFusion \cite{dai2017bundlefusion} that leverage more memory-efficient volumetric representations. While these systems perform online depth fusion, they usually require offline MarchingCubes \cite{Lorensen1987} to output final mesh models; \cite{Klingensmith2015,Steinbrucker2014,Wei2018} incorporates online meshing modules in such systems. 
Instead of utilizing volumetric spatial representations, \cite{Keller2013} proposes point-based fusion that maintains light-weight dense point cloud or surfels as 3D maps. ElasticFusion~\cite{Whelan2015} and InfiniTAM\_v3~\cite{InfiniTAM_V3_Report_2017} are efficient implementations of~\cite{Keller2013} with several extensions; \cite{Lefloch2017} further improves~\cite{Keller2013} by introducing surface curvatures. The point-based methods are unable to output mesh online, hence may not be suitable for physics-based applications.
\cite{Zhou2013,Choi_2015_CVPR} split scenes into fragments and register partially reconstructed mesh to build comprehensive models, but their offline property limits their usages.

\noindent \textbf{3D scene representations.} 
Dense 3D map requires efficient data structures to support high resolution reconstructions. For volumetric systems, plain 3D arrays \cite{Newcombe2011} are unsuitable for large scale scenes due to spatial redundancies. In view of this, moving volumes method \cite{Whelan2012Kintinuous} is introduced to maintain spatial properties only in active areas. Octree is used to ensure a complete yet adaptive coverage of model points \cite{Hornung:2013gt,Zeng2013,Steinbrucker2014}. As the tree might be unbalanced causing long traversing time, hierarchical spatial hashing is utilized \cite{Niessner2013,kahler2015very} supporting O(1) 3D indexing, and is further extended to be adaptive to local surface complexities~\cite{kahler2016hierarchical}.

There are also studies that directly represent scenes as point clouds or mesh during reconstruction. In \cite{Keller2013,Whelan2015} point clouds or surfels are simply arranged in an 1D array. Considering topologies in mesh, \cite{Marton2009} manages point clouds with inefficient KD-Tress for spatial resampling. \cite{Zienkiewicz2016} maintains a 2.5D map with fixed structured triangles which will fail to capture occlusions.
Hybrid data structures are also used to combine volumes and mesh. \cite{Steinbrucker2014} builds an octree-based structure where boundary conditions have to be carefully considered in term of mesh triangles. \cite{Klingensmith2015} uses spatial hashed blocks and stores mesh triangles in the block level, but ignores vertex sharing between triangles. \cite{Wei2018} reveals the correspondences between mesh vertices and voxel edges, reducing the redundancy in the aspect of data structure. Yet improvement is required to remove false surfaces generated from noisy input data.

\noindent \textbf{Uncertainty-aware data fusion.}
Uncertainty is one of the core problems remain in 3D reconstruction, which may come from imperfect inputs or complex environments. In volumetric representations that split the space into grids, probability distributions are usually utilized to model spatial properties. 
Binary occupancy is an intuitive variable configuration, denoting whether a voxel is physically occupied. \cite{Woodford:2012ee} proposes a joint distribution of point occupancy state and inlier ratio over the entire volume with visual constraints and achieves competitive results. \cite{ulusoy2015towards,Ulusoy:CVPR:2016} similarly emphasize ray-consistency to reconstruct global-consistent surfaces, whose inferences are too sophisticated to run in real-time. Although surface extraction can be performed on occupancy grids via thresholding or ray-casting, it is usually very sensitive to parameters. Instead of maintaining a $\{0, 1\}$ field, \cite{Curless1996} introduces SDF which holds signed projective distances from voxels to their closest surfaces sampled by input depth measurements. \cite{Newcombe2011,newcombe2012dense} use weight average of Truncated SDF in the data fusion stage by considering a per-voxel Gaussian distribution regarding SDF as a random variable. While Gaussian noise can be smoothed by weight average, outliers have to be carefully filtered out with ad hoc operations. \cite{Niessner2013} uses a temporal recycling strategy by periodically subtracting weight in volumes; \cite{Klingensmith2015} directly carves out noisy inputs; \cite{dai2017bundlefusion} proposes a weighted subtraction to de-integrate data which are assumed to be incorrectly registered. As a non-local prior, \cite{Dzitsiuk2017} refines SDF value on-the-go using plane fitting, which performs well mainly in flat scenes with relatively low voxel resolutions. We find a lack of systematic probabilistic solution for SDF dealing both Gaussian noise and possible outliers.

For point-based representations, \cite{vogiatzis2011video} proposes an elegant math model treating inlier ratio and depth of a point as random variables subject to a special distribution. The parameters of such distributions are updated in a Bayesian fashion. This approach is adopted by \cite{forster2014svo} in SLAM systems for inverse depths, achieving competitive results. 
An alternative is to select ad hoc weights involving geometry and photometric properties of estimated points and computing weighted average \cite{tanskanen2013live,Kolev2014}. This simple strategy shares some similarity to the fusion of SDF values, and is also used in RGB-D systems where depths are more reliable \cite{Keller2013,InfiniTAM_V3_Report_2017}. Despite the solid math formulations, point-based methods are comparatively prone to noise due to their discrete representations.
\section{Overview}
Our framework is based on the \textit{hybrid data structure} involving three \textit{3D representations} linked by \textit{PSDF}. The pipeline consists of iterative operations of data fusion and surface generation.

\begin{figure}[h]
	\centering
	\subfigure[Data structure] {
	\includegraphics[height=1.5in]{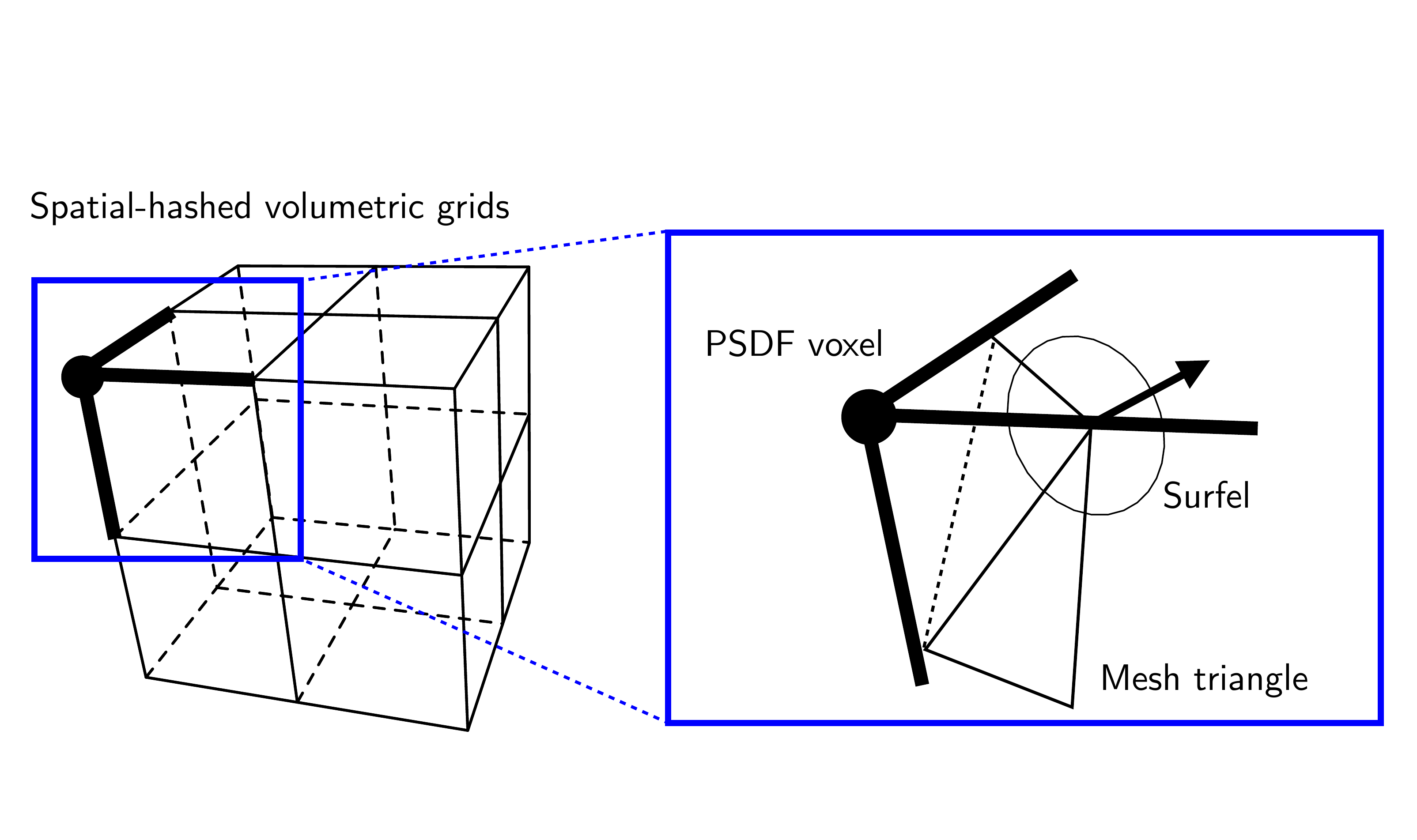}
	\label{fig:data-structure}
	}
	\subfigure[Pipeline] {
	\includegraphics[height=1.5in]{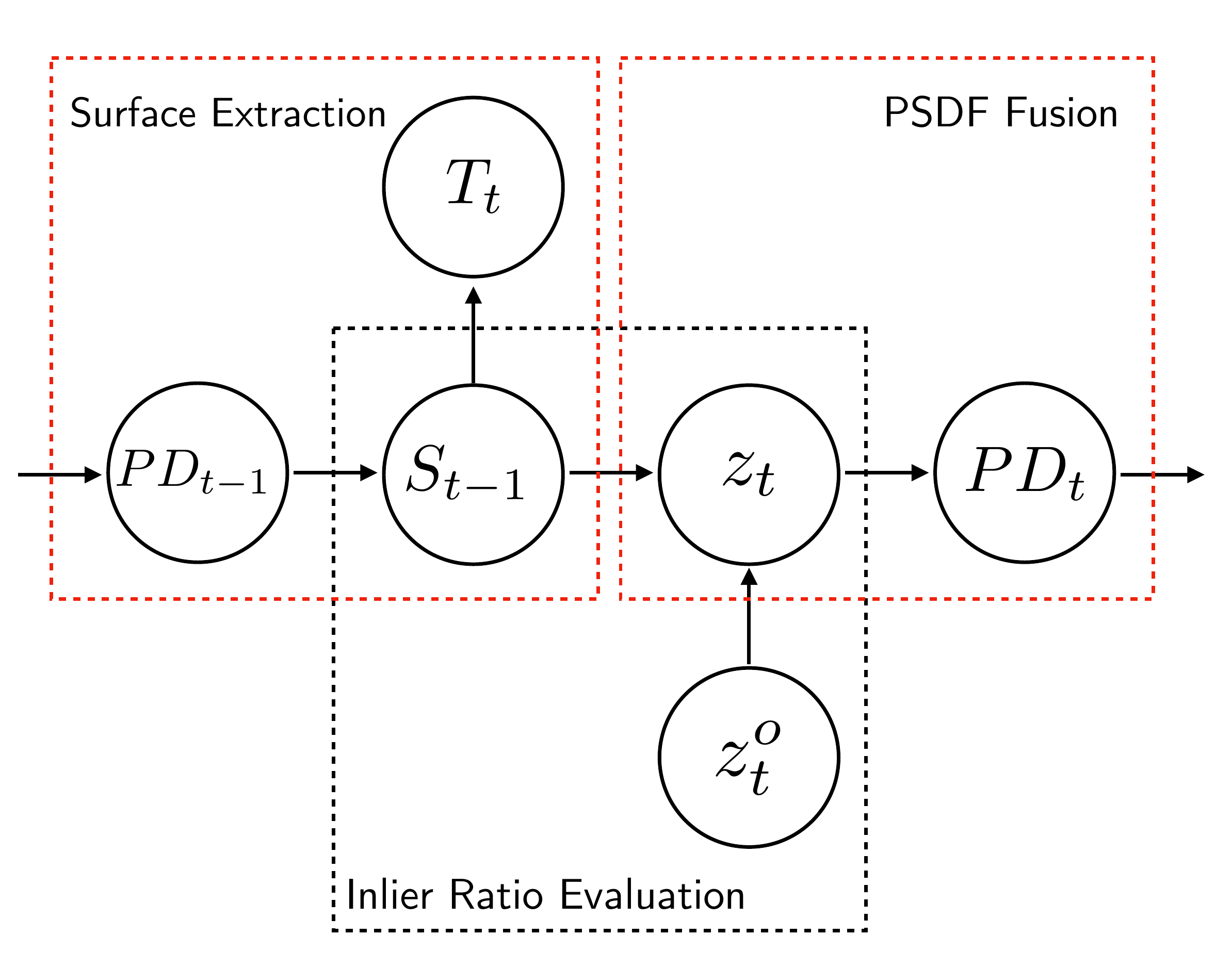}
	\label{fig:pipeline}
    }
	\caption{A basic hybrid unit, and the system pipeline.}\label{fig:pipeline+data-structure}
\end{figure}

\subsection{Hybrid Data Structure}
We follow \cite{Niessner2013,Klingensmith2015,Wei2018} and use a spatial hashing based structure to efficiently manage the space. A hash entry would point to a block, which is the smallest unit to allocate and free. A block is further divided into $8\times8\times8$ small voxels. Following \cite{Wei2018} we consider a voxel as a 3-edge structure instead of merely a cube, as depicted in Fig.\ref{fig:data-structure}, which will avoid ambiguity when we refer to shared edges. PSDF values are stored at the corners of these structures. In addition, we maintain surfels on the volumetric grids by limiting their degree of freedom on the edges of voxels; within a voxel at most 3 surfels on edges could be allocated. This constraint would regularize the distribution of surfels, guarantee easier access, and avoid duplicate allocation. Triangles are loosely organized in the level of blocks, linking adjacent surfels. In the context of mesh, a surfel could be also interpreted as a triangle vertex. 

\subsection{3D Representations}
\noindent \textbf{Voxel and PSDF}. In most volumetric reconstruction systems SDF or truncated SDF (TSDF) (denoted by $D$) of a voxel is updated when observed 3D points fall in its neighbor region. Projective signed distances from measurements, which could be explained as SDF observations, are integrated by computing weight average. Newcombe \cite{newcombe2012dense} suggests that it can be regarded as the solution of a maximum likelihood estimate of a joint Gaussian distribution taking SDF as a random variable. While Gaussian distribution could depict the uncertainty of data noise, it might fail to handle outlier inputs which are common in reconstruction tasks using consumer-level sensors. Moreover, SDF should depict the projective distance from a voxel to its \textit{closest} surface. During integration, however, it is likely that \textit{non-closest} surface points are taken into account, which should also be regarded as outlier SDF observations. In view of this, we introduce another random variable $\pi$ to denote the inlier ratio of SDF, initially used in \cite{vogiatzis2011video} to model the inlier ratio of 3D points:
\begin{align}
p(D_i^o \mid D, \tau_i, \pi) = \pi\mathcal{N}(D_i^o; D, \tau_i^2) + (1-\pi)\mathcal{U}(D_i^o; D_{min}, D_{max}), \label{equ:rvd}
\end{align}
where $D_i^o$ reads an SDF observation computed with depth measurements, $\tau_i$ is the variance of the SDF observation, $\mathcal{N}$ and $\mathcal{U}$ are Gaussian and Uniform distributions.

Following \cite{vogiatzis2011video}, the posterior of PSDF can be parameterized by a Beta distribution multiplying a Gaussian distribution $\mathcal{B}(a, b)\mathcal{N}(\mu; \sigma^2)$, given a series of observed input SDF measurements. The details will be discussed in \S\ref{subsec:bayesian}. The parameters $a, b, \mu, \sigma$ of the parameterized distribution are maintained per voxel.

\noindent \textbf{Surfel}. A surfel in our pipeline is formally defined by a position $\vec{x}$, a normal $\vec{n}$, and a radius $r$. Since a certain surfel is constrained on an edge in the volume, $\vec{x}$ is generally an interpolation of 2 adjacent voxel corners.

\noindent \textbf{Triangle}. A triangle consists of 3 edges, each linking two adjacent surfels. These surfels can be located in different voxels, even different blocks. In our framework triangles are mainly extracted for rendering; the contained topology information may be further utilized in extensions.

\noindent \textbf{Depth input}. We receive depth measurements from sensors as input, while sensor poses are assumed known. Each observed input depth $z^o$ is modeled as a random variable subject to a simple Gaussian distribution:
\begin{align}
p(z^o \mid z, \iota) = \mathcal{N}(z^o; z, \iota^2),\label{equ:rvz}
\end{align}
where $\iota$ can be estimated from a precomputed sensor error model.

\subsection{Pipeline}
In general, our system will first generate a set of surfels $S_{t-1}$ in the volumetric PSDF field $PD_{t-1}$. Meanwhile, mesh triangle set $T_t$ is also determined by linking reliable surfels in $S_{t-1}$. $S_{t-1}$ explicitly defines the surfaces of the scene, hence can be treated as a trustworthy geometry cue to estimate outlier ratio of the input depth data $z_t^o$.  $PD_{t-1}$ is then updated to $PD_t$ by fusing evaluated depth data distribution $z_t$ via Bayesian updating. The process will be performed iteratively every time input data come, as depicted in Fig.\ref{fig:pipeline}. We assume the poses of the sensors are known and all the computations are in the world coordinate system. 
\section{PSDF Fusion and Surface Reconstruction}

\begin{figure}[t]
	\centering
	\subfigure[PSDF update] {		
		\includegraphics[height=1.2in]{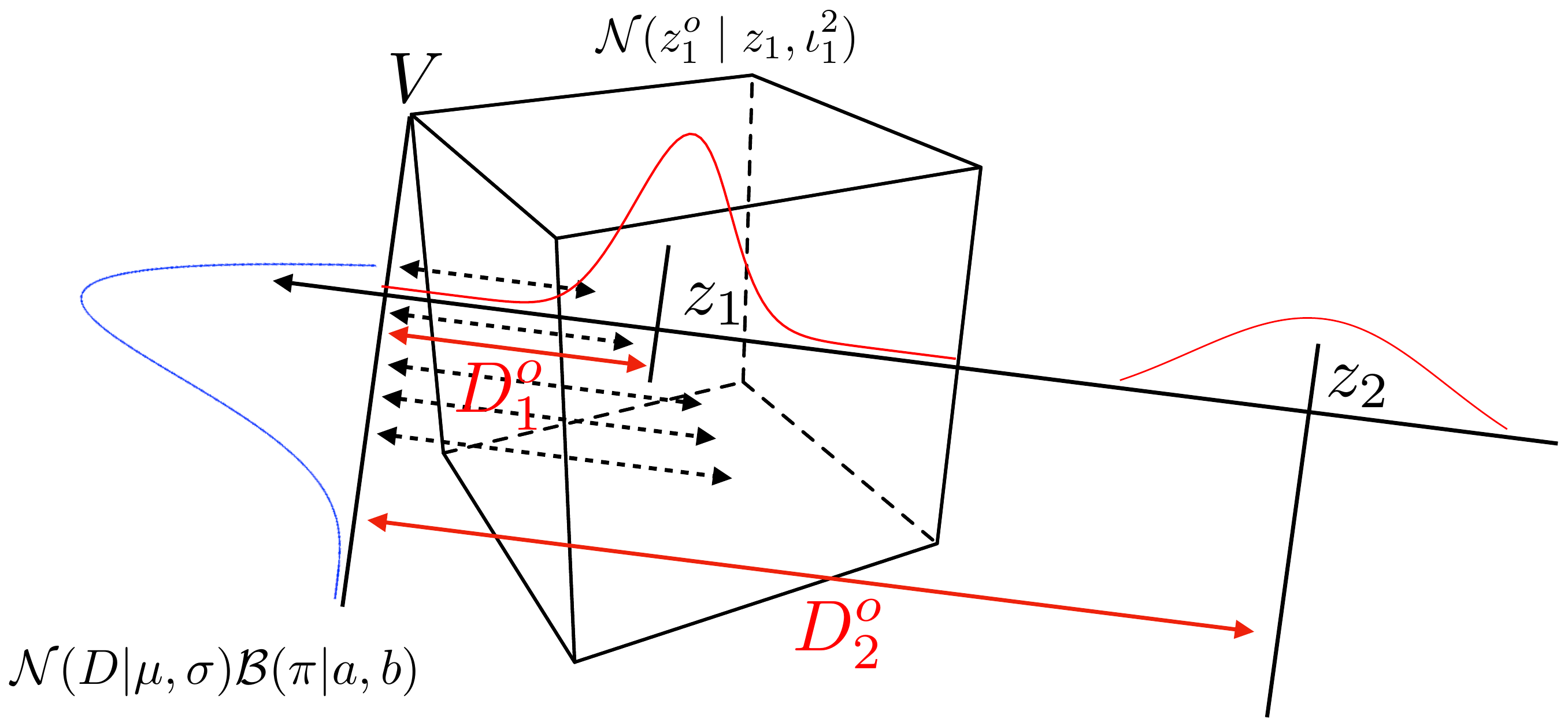} \label{fig:fusion}
	}
	\subfigure[Inlier ratio estimation] {
		\includegraphics[height=1.0in]{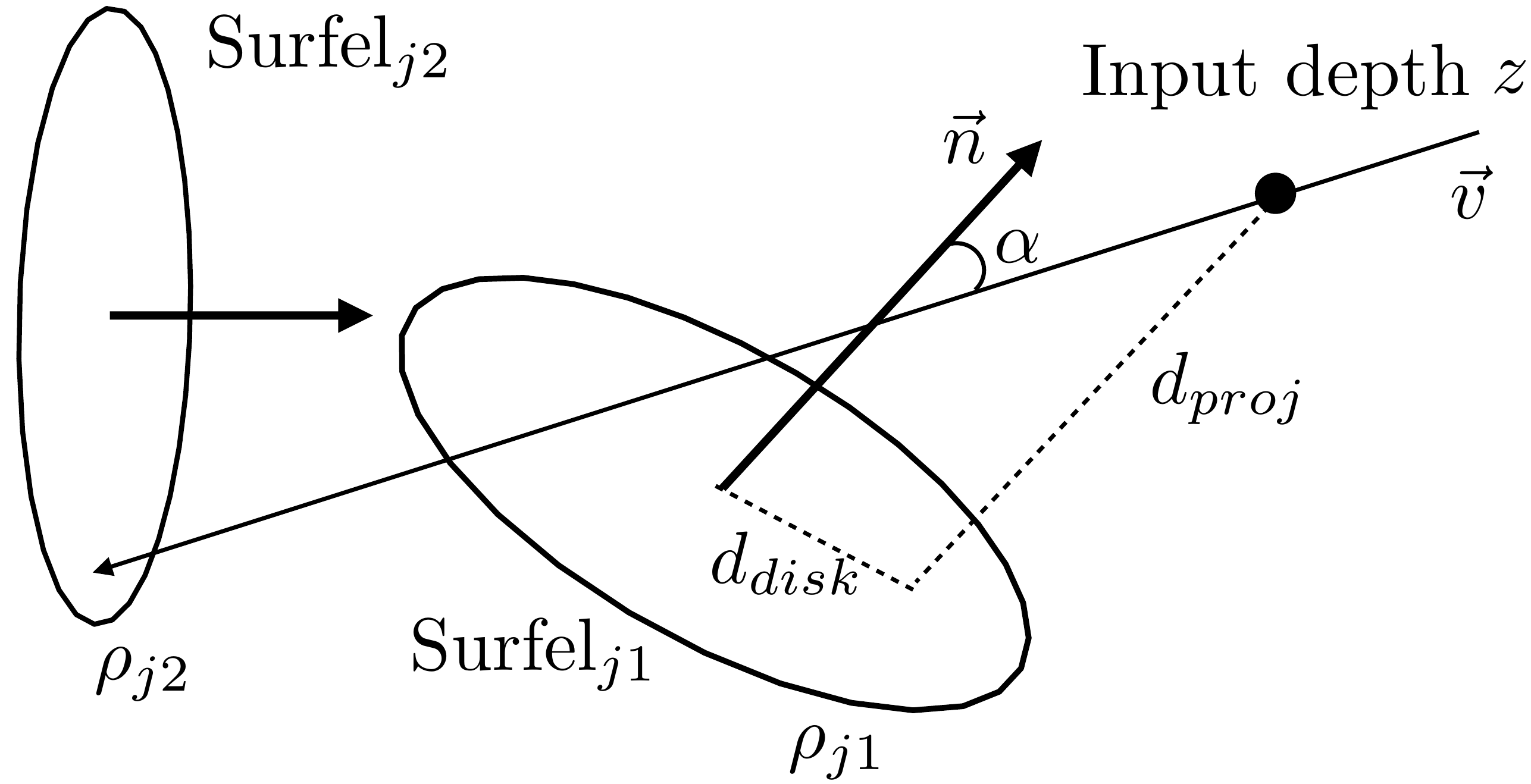} \label{fig:prediction}
	}
	\caption{Left, illustration of PSDF distribution with multiple SDF observations ($D_1^o$ is likely to be an inlier observation, $D_2^o$ is possibly an outlier). The red curves show Gaussian distributions per observation. The blue curve depicts a Beta distribution which intuitively suggests that inlier observations should be around $D_1^o$. Right, estimation of SDF inlier ratio $\rho$ involving observed input 3D point and already known surfels.}\label{fig:prediction+fusion}
\end{figure}

\subsection{PSDF Fusion}\label{subsec:bayesian}
Similar to \cite{Newcombe2011}, in order to get SDF observations of a voxel $V$ given input 3D points from depth images, we first project $V$ to the depth image to find the projective closest depth measurement $z_i$. Signed distance from $V$ to the input 3D data is defined by
\begin{align}
D_i = z_i - z^V \label{equ:sdf},
\end{align}
where $z^V$ is a constant value, the projective depth of $V$ along the scanning ray. 

The observed $D_i^o$ is affected by the variance $\iota_i$ of $z_i$ in Eq.\ref{equ:rvz} contributing to the Gaussian distribution component in Eq.\ref{equ:rvd}, provided $D_i$ is an inlier. Otherwise, $D_i$ would be counted in the uniform distribution part. Fig.\ref{fig:fusion} illustrates the possible observations of SDF in one voxel. 

Variance $\iota_i$ can be directly estimated by pre-measured sensor priors such as proposed in \cite{Nguyen:2012fq}. In this case, due to the simple linear form in Eq.\ref{equ:sdf}, we can directly set $D_i^o = z_i^o - z^V$ and $\tau_i = \iota_i$ in Eq.\ref{equ:rvd}. 

Given a series of independent observations $D_i^o$, we can derive the posterior
\begin{align}
p(D, \pi \mid D_1^o,\tau_1 \cdots D_n^o, \tau_n) 
\propto p(D, \pi) \prod_i p(D_i^o \mid D, \tau_i^2, \pi),
\end{align}
where $p(D, \pi)$ is a prior and $p(D_i^o \mid D, \tau_i^2, \pi)$ is defined by Eq.\ref{equ:rvd}.
It would be intractable to evaluate the production of such distributions with additions. Fortunately, \cite{vogiatzis2011video} proved that the posterior could be approximated by a parametric joint distribution:
\begin{align}
&p(D, \pi \mid D_1^o,\tau_1 \cdots D_n^o, \tau_n) 
\propto
\mathcal{B}(\pi \mid a_n, b_n)\mathcal{N}(D \mid \mu_n, \sigma_n^2), 
\end{align}
therefore the problem could be simplified as a parameter estimation in an incremental fashion:
\begin{align}
&\mathcal{B}(\pi \mid a_n, b_n)\mathcal{N}(D \mid \mu_n, \sigma_n^2) \propto
p(D_n^o | D, \tau_n^2, \pi)
\mathcal{B}(\pi \mid a_{n-1}, b_{n-1})\mathcal{N}(D \mid \mu_{n-1}, \sigma_{n-1}^2).
\end{align}
In \cite{vogiatzis2011video} by equating first and second moments of the random variables $\pi$ and $D$, the parameters could be easily updated, in our case evoking the change of SDF distribution and its inlier probability:
\begin{align}
\mu_n &= \frac{C_1}{C_1+C_2} m + \frac{C_2}{C_1+C_2} \mu_{n-1},\\
\mu_n^2 + \sigma_n^2 &= \frac{C_1}{C_1+C_2} (s^2 + m^2) + \frac{C_2}{C_1+C_2} (\sigma_{n-1}^2 + \mu_{n-1}^2), \\
s^2 &= 1 / (\frac{1}{\sigma_{n-1}^2} + \frac{1}{\tau_n^2}),\\
m &= s^2({\frac{\mu_{n-1}}{\sigma_{n-1}^2} + \frac{D_i^o}{\tau_n^2}}),\\
C_1 &= \frac{a_{n-1}}{a_{n-1}+b_{n-1}} \mathcal{N}(D_i^o ; \mu_{n-1},\sigma_{n-1}^2 + \tau_n^2), \label{equ:c1}\\
C_2 &= \frac{b_{n-1}}{a_{n-1}+b_{n-1}} \mathcal{U}(D_i^o; D_{min}, D_{max}),\label{equ:c2}
\end{align}
the computation of $a$ and $b$ are the same as \cite{vogiatzis2011video} hence ignored here. In our experiments we find that a truncated $D_i^o$ leads to better results, as it directly rejects distant outliers. SDF observations from non-closest surfaces are left to be handled by PSDF.

\subsection{Inlier Ratio Evaluation}\label{subsec:prediction}
In Eq.\ref{equ:c1}-\ref{equ:c2}, the expectation of $\mathcal{B}(\pi \mid a, b)$ is used to update the coefficients, failing to make full use of known geometry properties in scenes. In our pipeline, available surface geometry is considered to evaluate the inlier ratio $\rho_n$ of $D_n^o$, replacing the simple $\frac{a_{n-1}}{a_{n-1}+b_{n-1}}$.
Note $\rho_n$ is computed per-frame in order to update $C_1,C_2$; $\pi$ is still parameterized by $a$ and $b$.

$\rho_n$ can be determined by whether an input point $z$ is near the closest surface of a voxel and results in an inlier SDF observation. We first cast the scanning ray into the volume and collect the surfels maintained on the voxels hit by the ray. Given the surfels, 3 heuristics are used, as illustrated in Fig.\ref{fig:prediction}.\\
\textbf{Projective distance}. This factor is used to measure whether a sampled point is close enough to a surfel which is assumed the nearest surface to the voxel:
\begin{align}
w_{dist} = \exp(\frac{- |\vec{n}^T(\vec{x} - z \vec{v})|^2}{2\theta^2}),
\end{align}
where $\vec{v}$ is the normalized direction of the ray in world coordinate system and $\theta$ is a preset parameter proportional to the voxel resolution. 

\noindent \textbf{Angle}. Apart from projective distance, we consider angle as another factor, delineating the possibility that a scanning ray will hit a surfel. We use the empirical angle weight in \cite{Kolev2014}:
\begin{align}
w_{angle} = 
\begin{cases}
&\frac{\cos (\alpha) - \cos(\alpha_{max})}{1 - \cos(\alpha_{max})}, \text{if~} \alpha < \alpha_{max},\\
&w_{angle}^0, \text{~~~~~~~~~~~~else},
\end{cases}
\end{align}
where $\alpha = \langle \vec{n}, \vec{v}\rangle$, $\alpha_{max}$ is set to 80 deg and $w_{angle}^0$ assigned to $0.1$.\\
\textbf{Radius}. The area that surfels could influence vary, due to the local shape of the surface. The further a point is away from the center of a surfel, the less possible it would be supported. A sigmoid-like function is used to encourage a smooth transition of the weight:
\begin{align}
w_{radius} &= \gamma + \frac{2 (1 - \gamma)}{1 + \exp(\frac{-d_{disk}}{r})},\\
d_{disk} &= \sqrt{(z \vec{v} - \vec{x})^T(I - \vec{n} \vec{n}^T)(z \vec{v} - \vec{x})},
\end{align}
where parameter $\gamma \in [0, 1)$ and is set to $0.5$ in our case.

Putting all the factors together, we now have
\begin{align}
\rho = w_{dist} \cdot w_{radius} \cdot w_{angle}.
\end{align}
To compute the $\rho$ predicted by all the surfels, one may consider either summations or multiplications. However, we choose the highest $\rho$ instead -- intuitively a depth measurement is a sample on a surface, corresponding to exactly one surfel. A more sophisticated selection might include a ray consistency evaluation \cite{Woodford:2012ee,Ulusoy:CVPR:2016,ulusoy2015towards} where occlusion is handled. When a new area is explored where no surfels have been extracted, we use a constant value $\rho_{pr} = 0.1$ to represent a simple occupancy prior in space, hence we have
\begin{align}
\rho = \max_j \{\rho_{pr}, \rho_{1}, \cdots, \rho_{j}, \cdots \}.
\end{align}

\subsection{Surface Extraction}
PSDF implicitly defines zero crossing surfaces and decides whether they are true surfaces. The surface extraction is divided into two steps.\\
\textbf{Surfel generation}. In this stage we enumerate zero-crossing points upon 3 edges of each voxel and generate surfels when condition
\begin{align}
\mu^{i1} \cdot \mu^{i2} &< 0, \nonumber\\
\frac{a^{i1}} {a^{i1} + b^{i1}} > \pi_{thr}& \text{~and~} \frac{a^{i2}} {a^{i2} + b^{i2}} > \pi_{thr},
\end{align}
are satisfied, where $i1$ and $i2$ are indices of adjacent voxels and $\pi_{thr}$ is a confidence threshold. Supported by the reliable update of the PSDF, false surfaces could be rejected and duplicates could be removed. According to our experiments, our framework is not sensitive to $\pi_{thr}$; $0.4$ would work for all the testing scenes.
A surfel's position $\vec{x}$ would be the linear interpolation of corresponding voxels' positions $\vec{x}^i$ indexed by $i$, and the radius would be determined by $\sigma$ of adjacent voxels, simulating its affecting area. Normal is set to normalized gradient of the SDF field, as mentioned in \cite{Newcombe2011}. 
\begin{align}
\vec{x} &= \frac{|\mu^{i2}|}{|\mu^{i1}| + |\mu^{i2}|} \vec{x^{i1}} + \frac{|\mu^{i1}|}{|\mu^{i1}| + |\mu^{i2}|} \vec{x^{i2}},\\
r &= \frac{|\mu^{i2}|}{|\mu^{i1}| + |\mu^{i2}|} \sigma^{i1} + \frac{|\mu^{i1}|}{|\mu^{i1}| + |\mu^{i2}|}\sigma^{i2},\\
\vec{n} &= \nabla \mu / || \nabla \mu ||.
\end{align}
\textbf{Triangle generation.}
Having sufficient geometry information within surfels, there is only one more step to go for rendering-ready mesh. The connections between adjacent surfels are determined by the classical MarchingCubes \cite{Lorensen1987} method. As a simple modification, we reject edges in the voxel whose $\sigma$ is larger than a preset parameter $\sigma_{thr}$. This operation will improve the visual quality of reconstructed model while preserving surfels for the prediction stage.
\begin{figure}[t]
	\centering
	\includegraphics[width=\textwidth]{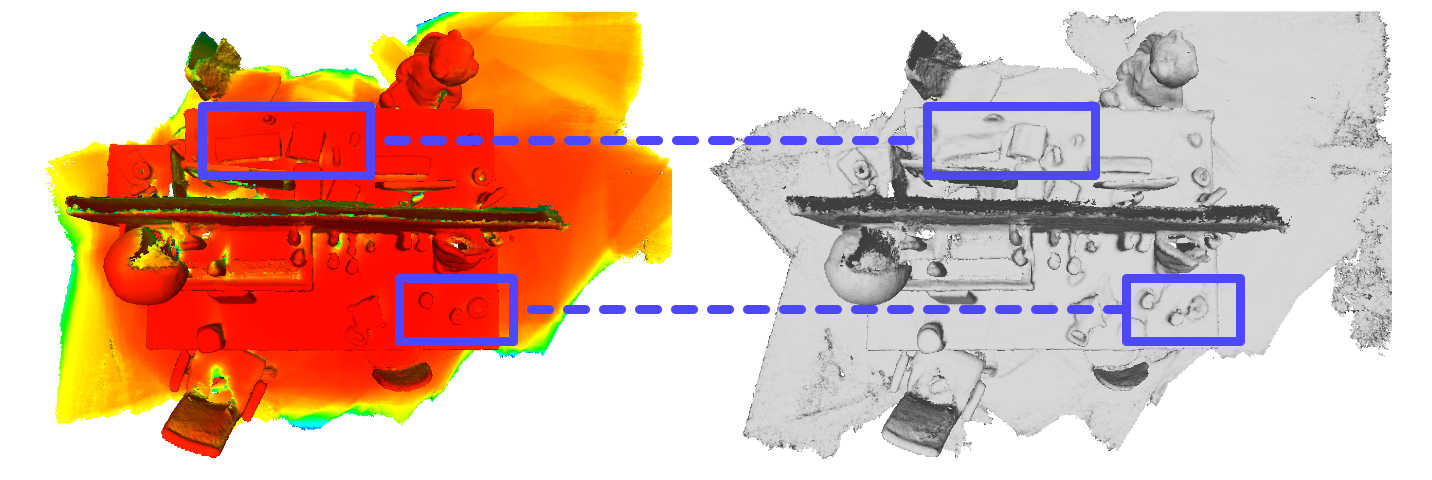}
	\caption{Comparison of output mesh of \textit{frei3\_long\_office} scene. Left, \textit{PSDF}. Right, \textit{TSDF}. Our method generates smooth surfaces and clean edges of objects, especially in blue rectangles.}\label{fig:tum3}
\end{figure}

\section{Experiments}
We test our framework (denoted by {\it PSDF}) on three RGB-D datasets: TUM \cite{sturm12iros}, ICL-NUIM \cite{handa2014}, and dataset from Zhou and Koltun \cite{Zhou2013}. Our method is compared against \cite{Wei2018} (denoted by \textit{TSDF}) which incrementally extracts mesh in spatial-hashed TSDF volumes. The sensors' poses are assumed known for these datasets, therefore the results of {\it TSDF} should be similar to other state-of-the-art methods such as \cite{Niessner2013,InfiniTAM_V3_Report_2017} where TSDF integration strategies are the same. 
We demonstrate that our method reconstructs high quality surfaces by both qualitative and quantitative results. Details are preserved while noise is removed in the output models. The running speed for online mesh extraction is also improved by avoiding computations on false surfel candidates.

For \cite{sturm12iros,Zhou2013} we choose a voxel size of 8mm and $\sigma_{thr} = $ 16mm; for \cite{handa2014} voxel size is set to 12mm and $\sigma_{thr} = $ 48mm. The truncation distance is set to $3\times$ voxel size plus $3\times\tau$; with a smaller truncation distance we found strides and holes in meshes. Kinect's error model \cite{Nguyen:2012fq} was adopted to get $\iota$ where the factor of angle was removed, which we think might cause double counting considering $w_{angle}$ in the inlier prediction stage. The program is written in C++/CUDA 8.0 and runs on a laptop with an Intel i7-6700 CPU and an NVIDIA 1070 graphics card.

\begin{figure}[t]
	\centering
	\includegraphics[width=\textwidth]{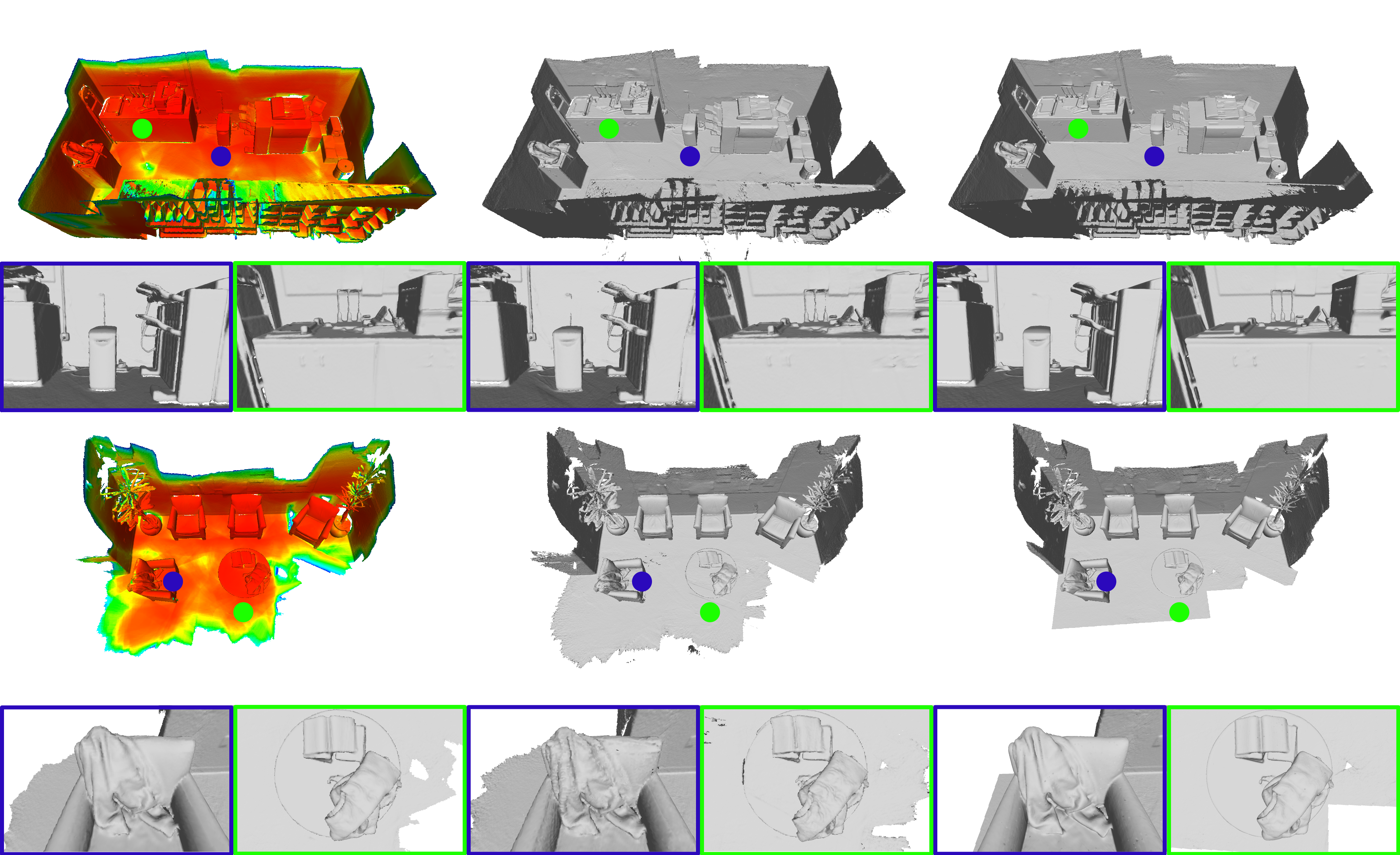}
	\caption{Output mesh of the {\it copyroom} and \textit{lounge} scenes. From left to right, \textit{PSDF}, \textit{TSDF}, mesh provided by \cite{Zhou2013}. Zoomed in regions show that our method is able to filter outliers while maintaining complete models and preserving details. Best viewed in color.}\label{fig:copyroom+lounge}
\end{figure}

\subsection{Qualitative Results}
We first show that \textit{PSDF} accompanied by the related mesh extraction algorithm produces higher quality surfaces than \textit{TSDF}. Our results are displayed with shaded heatmap whose color indicates the inlier ratio of related SDF. Both geometry and probability properties can be viewed in such a representation.

Fig.\ref{fig:tum3} shows that \textit{PSDF} outperforms \textit{TSDF} by generating clean boundaries of small objects and rejecting noisy areas on the ground. In Fig.\ref{fig:copyroom+lounge}, in addition to the results of \textit{TSDF}, we also display the reconstructed mesh from offline methods provided by \cite{Zhou2013} as references. It appears that our method produces results very similar to \cite{Zhou2013}. While guaranteeing well-covered reconstruction of scenes, we filter outliers and preserve details. In \textit{copyroom}, the wires are completely reconstructed, one of which above PC is smoothed out in \cite{Zhou2013} and only partially recovered by \textit{TSDF}. In \textit{lounge}, we can observe a complete shape of table, and de-noised details in the clothes.

\begin{figure}[t]
	\centering
	\includegraphics[width=1.0\textwidth]{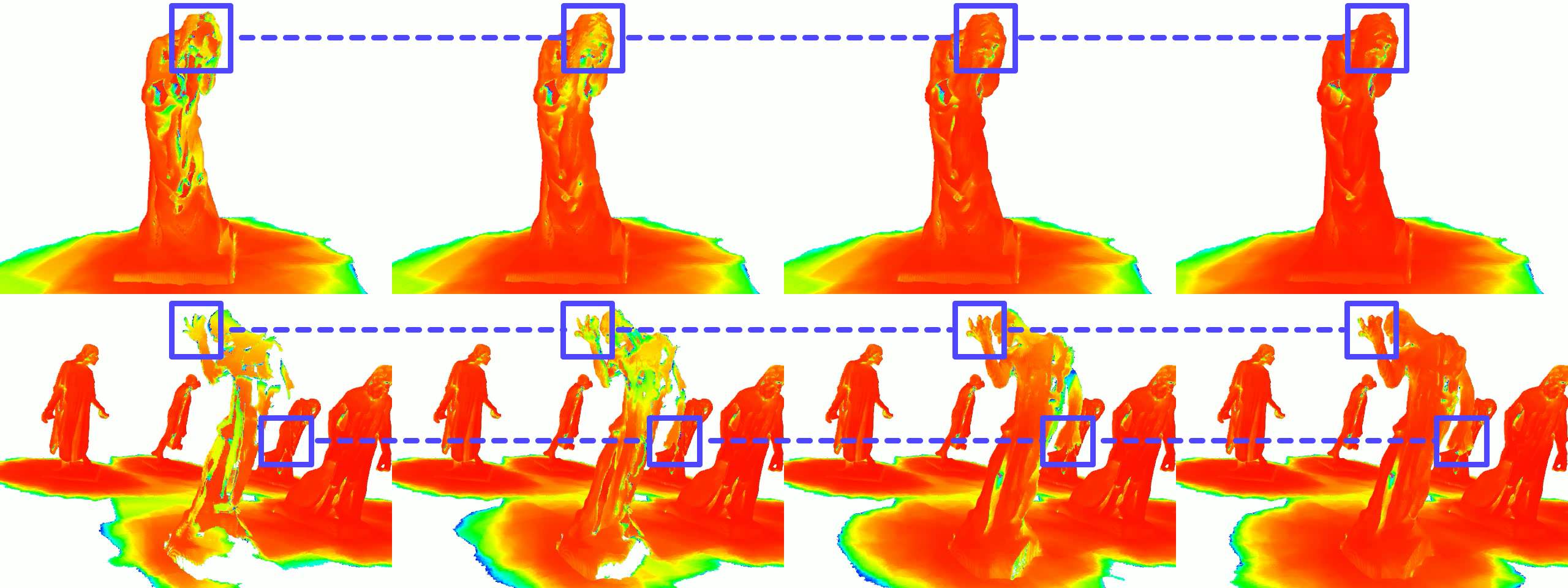}
	\caption{Incremental reconstruction of the {\it burghers} dataset. Fluctuation of probability could be observed at error-prone regions as an indication of uncertainty propagation.}\label{fig:burghers-sequential}
\end{figure}

We also visualize the incremental update of $\pi$ by rendering $\mathbb{E}(\beta(\pi | a, b)) = a/(a+b)$ as the inlier ratio of reconstructed mesh in sequence using colored heatmap. Fig.\ref{fig:burghers-sequential} shows the fluctuation of confidence around surfaces. The complex regions such as fingers and wrinkles on statues are more prone to noise, therefore apparent change of shape along with color can be observed.

\subsection{Quantitative Results}
\textbf{Reconstruction Accuracy}. We reconstruct mesh of the synthetic dataset \textit{livingroom2} with added noise whose error model is presented in \cite{handa2014}. Gaussian noise on inverse depth plus local offsets is too complicated for our error model, therefore we simplify it by assigning inverse sigma at certain inverse depths to $\iota_i$. The mesh vertices are compared with the ground truth point cloud using the free software \textit{CloudCompare} \cite{Cloudcompare}.

\begin{table}[h]
	\centering
	\begin{tabular}{P{2cm}P{4cm}P{4cm}}
		\specialrule{1pt}{1pt}{3pt}
		METHOD & {MEAN} (m) & {STD} (m)\\
		\midrule
		\textit{PSDF} & \textbf{0.011692} & \textbf{0.015702}\\
		\textit{TSDF} & 0.022556 & 0.076120\\
		\specialrule{1pt}{1pt}{3pt}
	\end{tabular}
	\caption{Statistics of the point-to-point distances from the reconstructed model vertices to the ground truth point cloud. \textit{PSDF} yields better reconstruction accuracy.} \label{table:cloudcompare}
\end{table}

\begin{figure*}[ht]
	\centering
	\subfigure[Point-to-point distance heatmap. Left, \textit{PSDF}. Right, \textit{TSDF}.] {
		\includegraphics[width=0.9\textwidth]{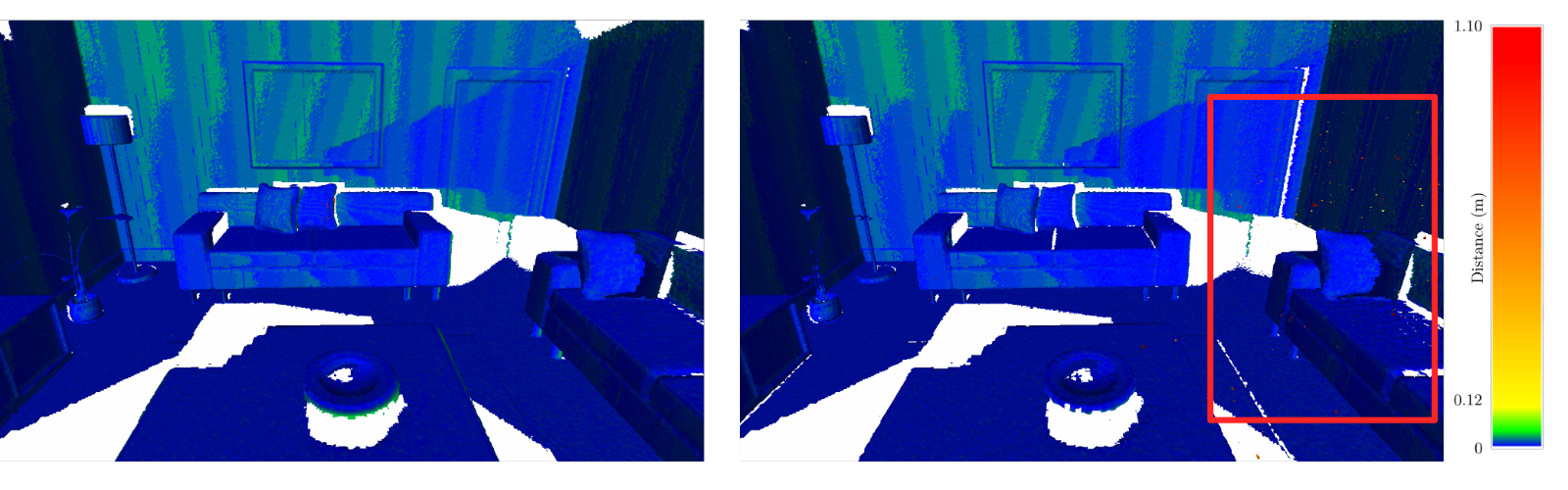}
	}
	\subfigure[Histogram head: $0$ to $4cm$] {
		\includegraphics[height=1.5in]{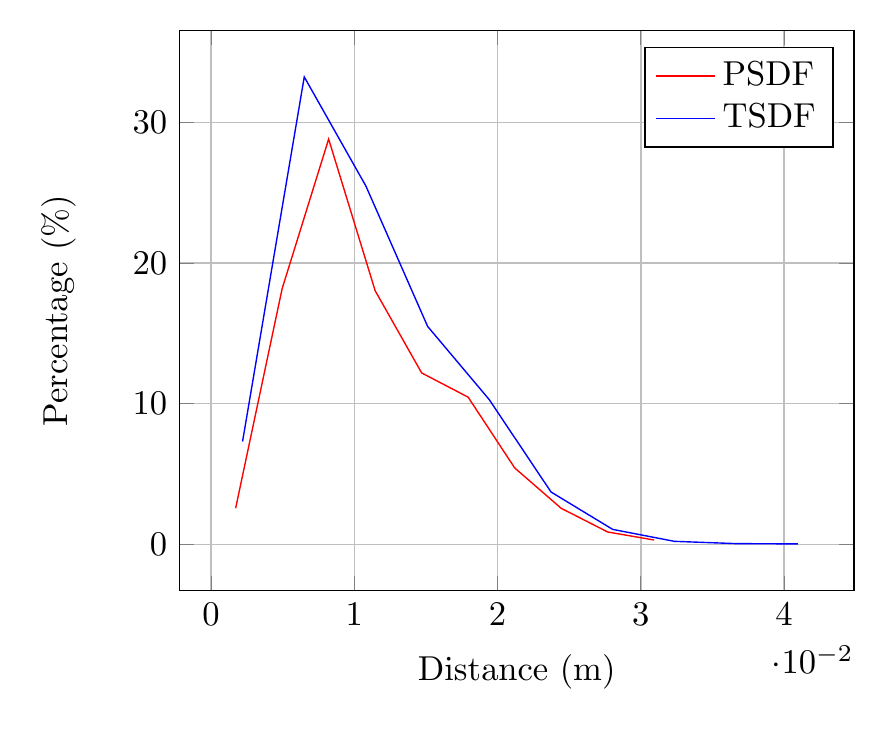}
	}
	\subfigure[Histogram tail: $4cm$ to $\infty$] {
		\includegraphics[height=1.5in]{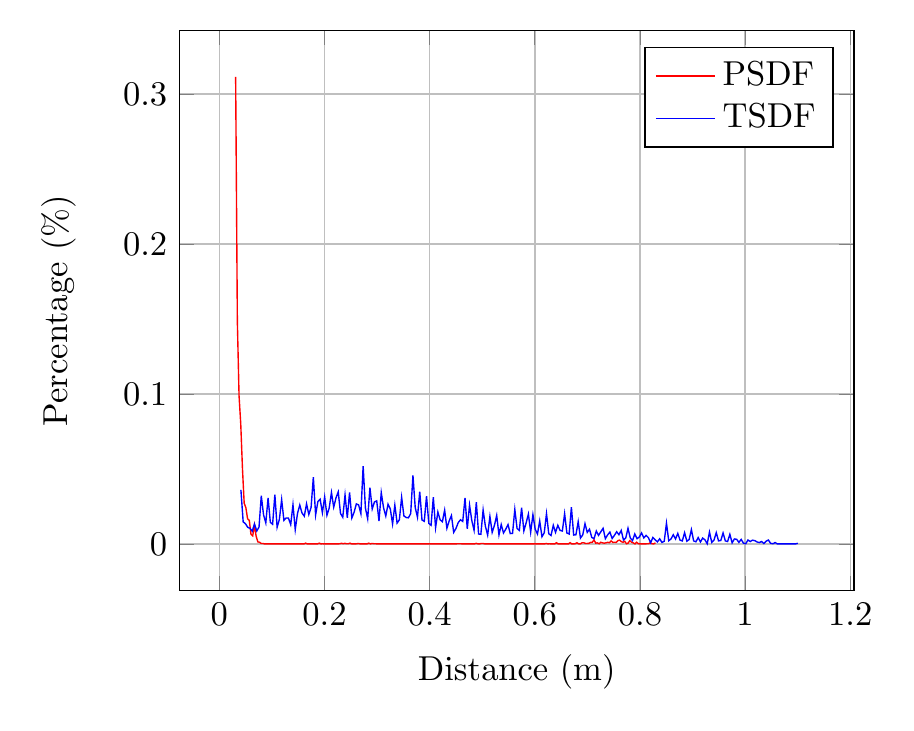}
	}
	\caption{Comparison of output mesh quality of {\it PSDF} and {\it TSDF}. First row: heatmap of the point-to-point distance from the reconstructed model to the ground truth. Notice the outliers in the red rectangle. Second row, the distance distribution to ground truth. The distribution is divided into two parts in order to emphasize the existence of outliers within the reconstruction of \textit{TSDF}, while \textit{PSDF} does not face such problems. Best viewed in color.}\label{fig:cloudcompare}
\end{figure*}

Table.\ref{table:cloudcompare} indicates that \textit{PSDF} reconstructs better models than \textit{TSDF}. Further details in Fig.\ref{fig:cloudcompare} suggest that less outliers appear in the model reconstructed by \textit{PSDF}, leading to cleaner surfaces.

\begin{figure}[h]
	\centering
	\subfigure[\textit{frei3\_long\_office}] {
		\includegraphics[height=1.2in]{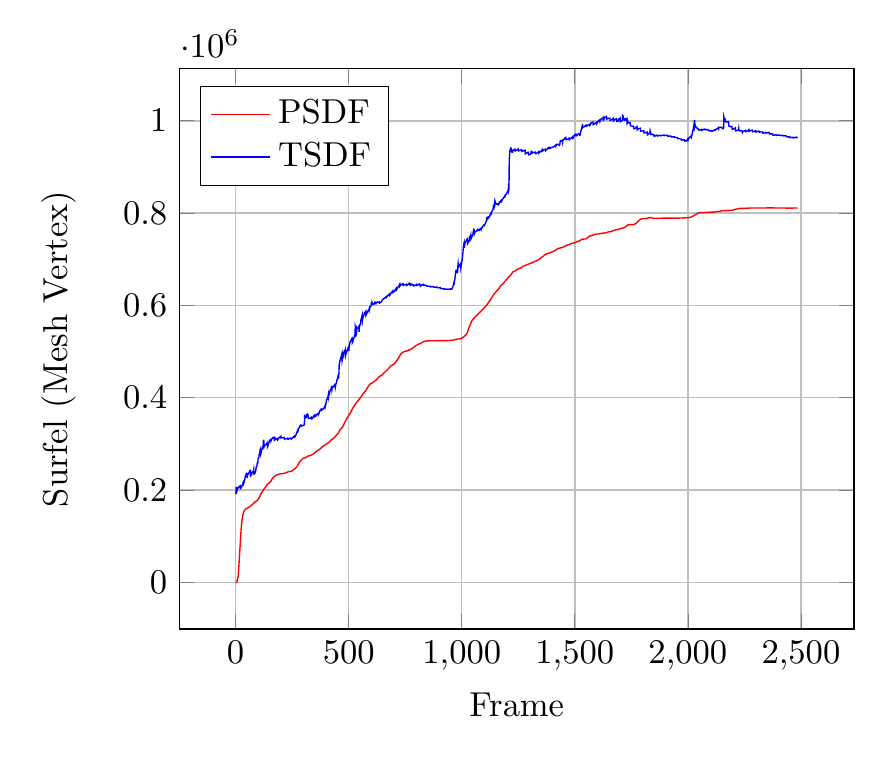}
	}
	\subfigure[\textit{lounge}] {
		\includegraphics[height=1.2in]{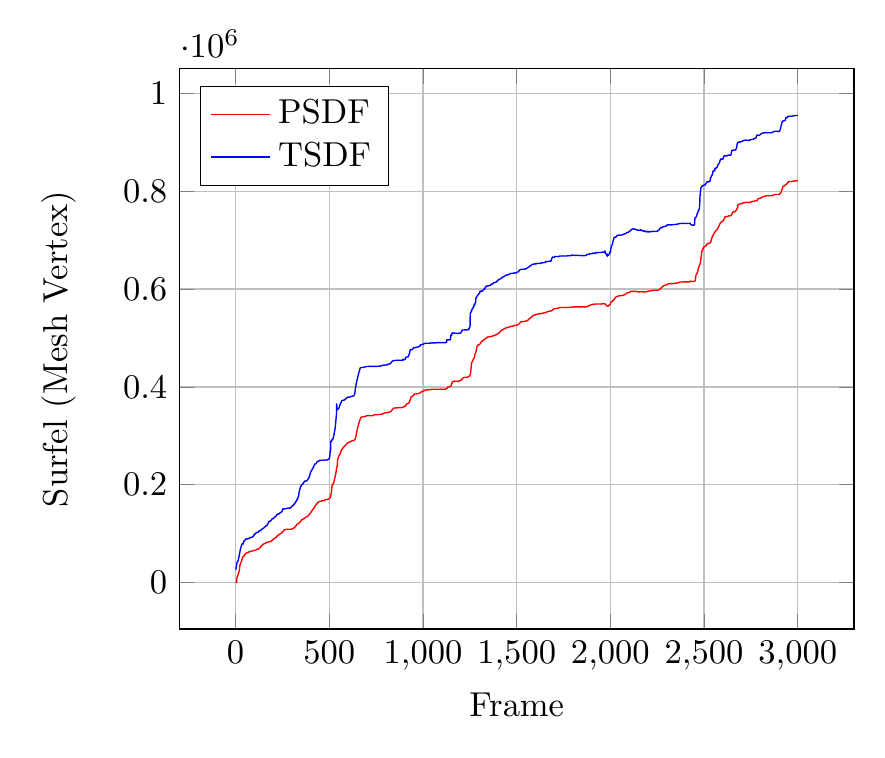}
	}
	\subfigure[Mapping] {
		\includegraphics[height=1.2in]{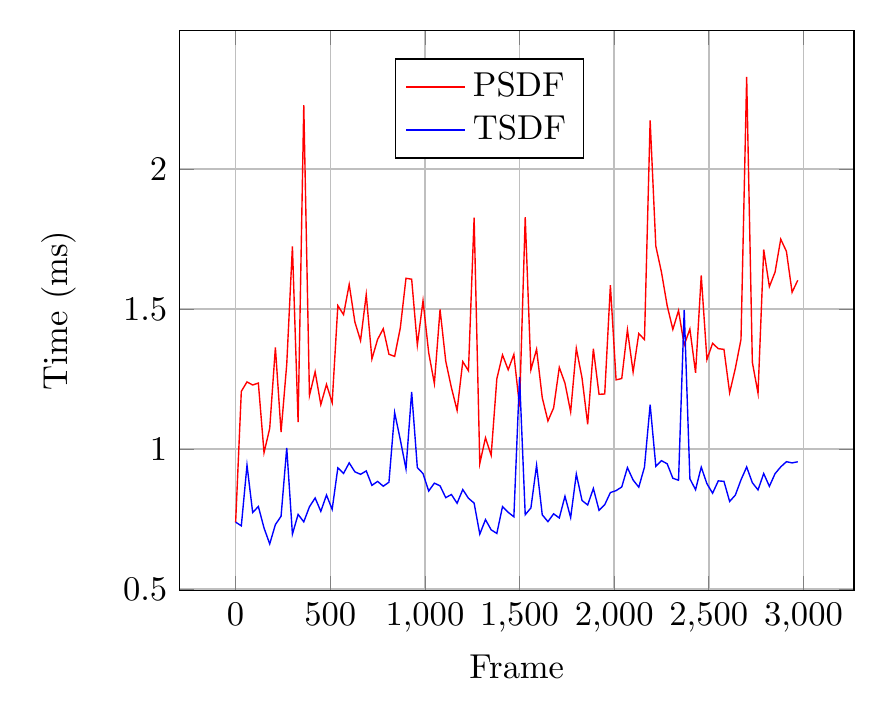}\label{fig:mapping-time}
	}
	\subfigure[Meshing] {
		\includegraphics[height=1.2in]{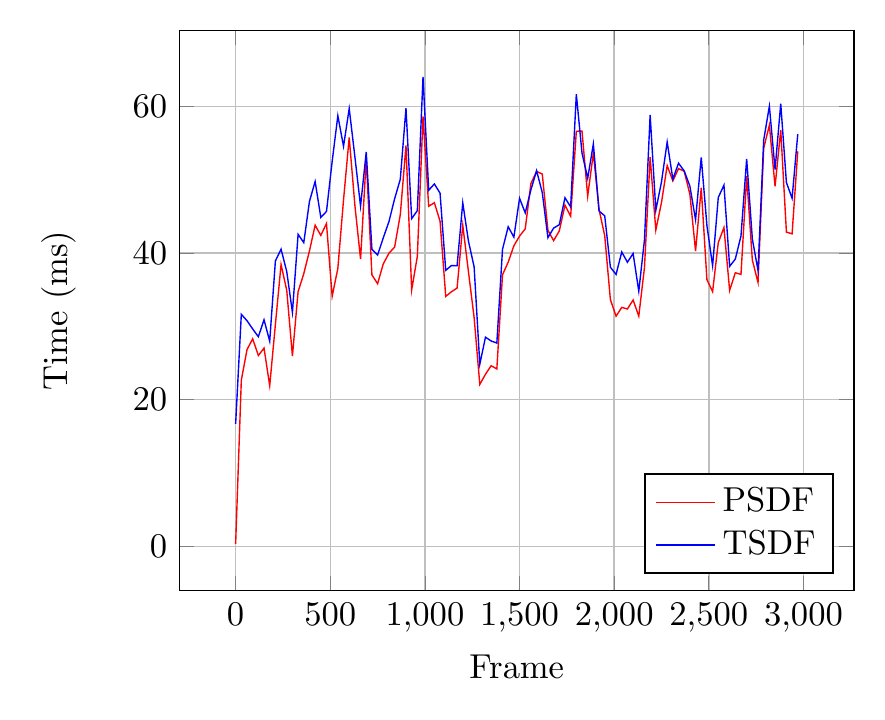}\label{fig:meshing-time}
	}
	\subfigure[All] {
		\includegraphics[height=1.2in]{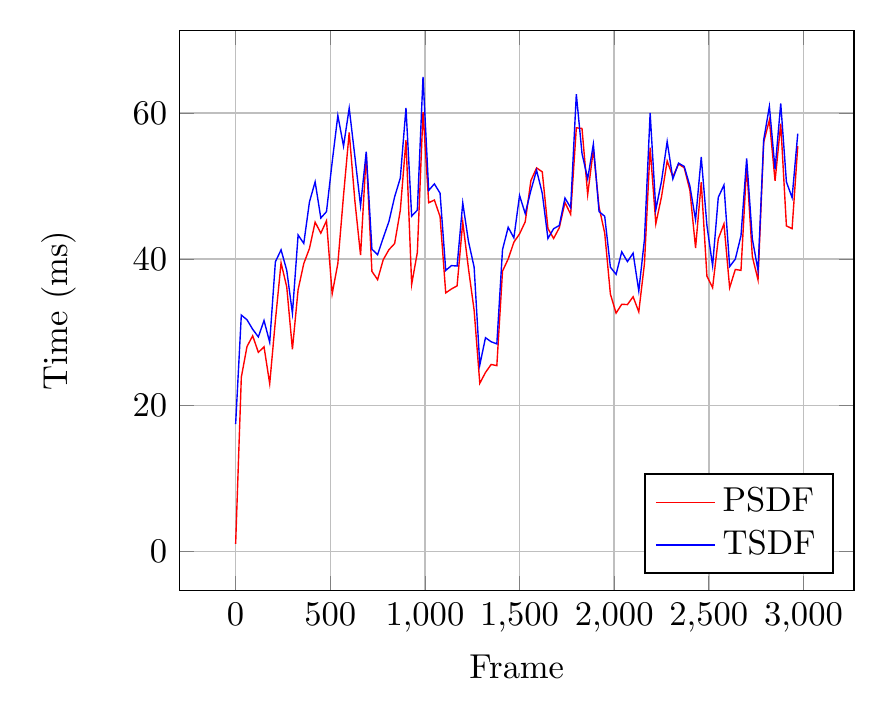}\label{fig:all-time}
	}
	\subfigure[Meshing vs raycasting] {
		\includegraphics[height=1.2in]{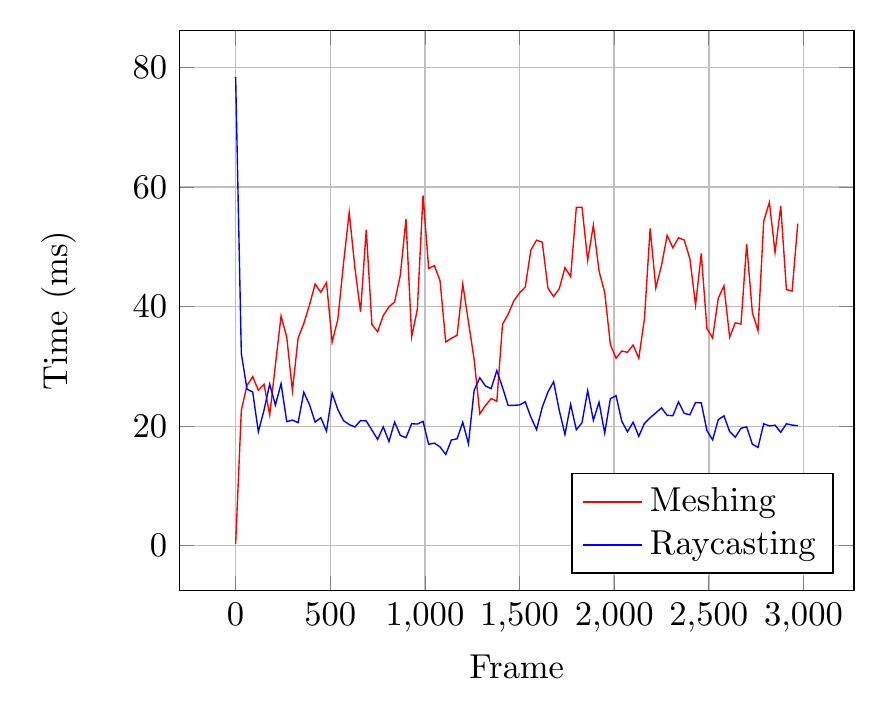}\label{fig:raycasting-time}
	}	
	\caption{(a)-(b), increasing trend of mesh vertices (surfels). (a), {\it frei3\_long\_office} scene in which a loop closure occurs at around 1500 frames. (b), {\it lounge} scene without loop closures. Mesh generated by \textit{PSDF} consumes approximately 80\% the memory of \textit{TSDF}. (c)-(f), running time analysis of our approaches and compared methods on \textit{lounge} scene. \textit{PSDF} is 1ms slower on GPU due to additional inlier prediction stage, but will save more time for surface extraction, causing a faster speed in general. The speed of online meshing is slower than raycasting but comparable.}\label{fig:time-mem}
\end{figure}

\noindent \textbf{Mesh size}. Our method maintains the simplicity of mesh by reducing false surface candidates caused by noise and outliers. As shown in Fig.\ref{fig:time-mem} and Table \ref{table:time-mem}, {\it PSDF} in most cases generates less vertices (\% 20) than {\it TSDF}, most of which are outliers and boundaries with low confidence. Fig.\ref{fig:time-mem} shows that the vertex count remains approximately constant when a loop closure occurred in {\it frei3\_long\_office}, while the increasing rate is strictly constrained in the {\it lounge} sequence where there is no loop closure.

\noindent \textbf{Time}. To make a running time analysis, we take real world {\it lounge} as a typical sequence where noise is common while the camera trajectory fits scanning behavior of humans. As we have discussed, evaluation of inlier ratio was performed, increasing total time of the fusion stage. However we find on a GPU, even with a relatively high resolution, the average increased time is at the scale of ms (see Fig.\ref{fig:mapping-time} and Table.\ref{table:time-mem}) and can be accepted. 

When we come to meshing, we find that by taking the advantage of PSDF fusion and inlier ratio evaluation, unnecessary computations can be avoided and {\it PSDF} method runs faster than {\it TSDF}, as plotted in Fig.\ref{fig:meshing-time}. The meshing stage is the runtime bottleneck of the approach, in general the saved time compensate for the cost in fusion stage, see Fig.\ref{fig:all-time} and Table.\ref{table:time-mem}.

We also compare the time of meshing to the widely used ray-casting that renders surfaces in real-time. According to Table \ref{table:time-mem}, in some scenes where sensor is close to the surfaces performing scanning, less blocks are allocated in viewing frustum and the meshing speed could be comparative to ray-casting, as illustrated in Fig.\ref{fig:raycasting-time}. As for other scenes requiring a large scanning range, especially {\it frei3\_long\_office} where more blocks in frustum have to be processed, ray-casting shows its advantage. We argue that in applications that only require visualization, ray-casting can be adopted; otherwise meshing offers more information and is still preferable.

\begin{table}
	\centering
	\resizebox{\textwidth}{!}{
		\begin{tabular}{|l|c|| *{2}{P{1.8cm}|} | *{4}{P{1.5cm}|} c || }
			\hline
			\multirow{3}{*}{Dataset} & \multirow{3}{*}{Frames} & \multicolumn{2}{c||}{MEMORY (vertex count)} & \multicolumn{5}{c||}{TIME (ms)} \\ 
			\cline{3-9}
			& & \multirow{2}{*}{PSDF} & \multirow{2}{*}{TSDF} & \multicolumn{2}{c|}{PSDF} & \multicolumn{2}{c|}{TSDF}  & \multirow{2}{*}{Ray-casting} \\ 
			\cline{5-8}
			& & & & Mapping & Meshing & Mapping & Meshing &\\ 
			\hline\hline
			\textit{burghers} & 11230 & \textbf{1362303} & {1438818} & {1.053} & {39.691} & \textbf{0.771} & \textbf{38.907} & {27.761} \\ 
			\textit{copyroom} & 5490 & \textbf{1222196} & {1328397} & {1.329} & \textbf{39.090} & \textbf{0.909} & {41.595} & {21.292} \\ 
			\textit{garden} & 6152 & \textbf{922534} & {978206} & {1.473} & {68.224} & \textbf{0.907} & \textbf{66.680} & {25.479} \\ 
			\textit{lounge} & 3000 & \textbf{821360} & {955218} & {1.331} & \textbf{41.270} & \textbf{0.881} & {45.248} & {22.097} \\ 
			\textit{livingroom1} & 965 & {529305} & \textbf{518885} & {1.255} & {33.090} & \textbf{0.743} & \textbf{32.865} & {27.248} \\ 
			\textit{livingroom2} & 880 & \textbf{609421} & {683008} & {1.407} & \textbf{33.446} & \textbf{0.759} & {40.230} & {29.069} \\ 
			\textit{office1} & 965 & \textbf{667614} & {674034} & {1.264} & \textbf{24.933} & \textbf{0.799} & {27.654} & {26.957} \\ 
			\textit{office2} & 880 & \textbf{712685} & {883138} & {1.322} & \textbf{35.029} & \textbf{0.767} & {45.923} & {29.981} \\ 
			\textit{frei1\_xyz} & 790 & \textbf{212840} & {352444} & {1.193} & \textbf{45.163} & \textbf{1.149} & {71.670} & {19.796} \\ 
			\textit{frei3\_long\_office} & 2486 & \textbf{811092} & {963875} & {2.424} & \textbf{159.417} & \textbf{1.375} & {161.485} & {26.545} \\ 
			\hline
		\end{tabular}
	}
\vspace*{3pt}
\caption{Evaluation of memory and time cost on various datasets. PSDF reduces model's redundancy by rejecting false surfaces and noise. The mapping stage of TSDF is faster, but in general PSDF spends less time considering both mapping and meshing stages.}\label{table:time-mem}
\end{table}

\section{Conclusions}
We propose PSDF, a joint probabilistic distribution to model the spatial uncertainties and 3D geometries. With the help of Bayesian updating, parameters of such a distribution could be incrementally estimated by measuring the quality of input depth data. Built upon a hybrid data structure, our framework can iteratively generate surfaces from the volumetric PSDF field and update PSDF values through reliable probabilistic data fusion supported by reconstructed surfaces. As an output, high-quality mesh can be generated on-the-fly in real-time with duplicates removed and noise cleared.

In the future we intend to extend our framework upon the basis of PSDF. We plan to establish a hierarchical data structure so that the resolution of spatial-hashed blocks could be adaptive to input data according to PSDF distributions. We also work on employing more priors to further enrich PSDF probability distributions. Localization modules using the proposed 3D representation are also planned to be integrated in our framework, in order to construct a complete probabilistic SLAM system.

%
%
%
\clearpage
\noindent \textbf{\large Acknowledgments}\\
This work is supported by the National Natural Science Foundation of China (61632003, 61771026), and National Key Research and Development Program of China (2017YFB1002601).

\bibliographystyle{splncs04}
\bibliography{egbib}
\end{document}